\documentclass[letterpaper, 10 pt, conference]{ieeeconf}  

\IEEEoverridecommandlockouts                              

\usepackage{graphics} 
\usepackage{epsfig} 
\usepackage{times} 
\usepackage{amsmath} 
\usepackage{amssymb}  
\usepackage{cuted}
\usepackage{capt-of}
\usepackage{float}
\usepackage{bm}
\usepackage{booktabs}
\usepackage{multirow}
\usepackage{microtype}
\usepackage{subcaption}
\usepackage{hyperref}
\usepackage{xcolor}
\hypersetup{
    hidelinks,
}

\title{\LARGE \bf
\textsc{BrickCraft}: Visuomotor Skill Composition with Situated Manual Guidance for Long-Horizon Interlocking Brick Assembly
}

\author{Jichuan Yu$^{*1,2}$, Bowei Li$^{*2}$, Zhenran Tang$^{2}$, Guanxing Lu$^{1,2}$, Chuxiong Hu$^{1}$, Ruixuan Liu$^{2}$ and Changliu Liu$^{2}$ 
\vspace{0.15cm}\\
$^{1}$ Tsinghua University \qquad $^{2}$ Carnegie Mellon University
\thanks{$^{*}$ Indicates equal contribution.}
}

\begin{document}


\maketitle
\thispagestyle{empty}
\pagestyle{empty}

\begin{strip}
	\centering
    \includegraphics[width=\linewidth]{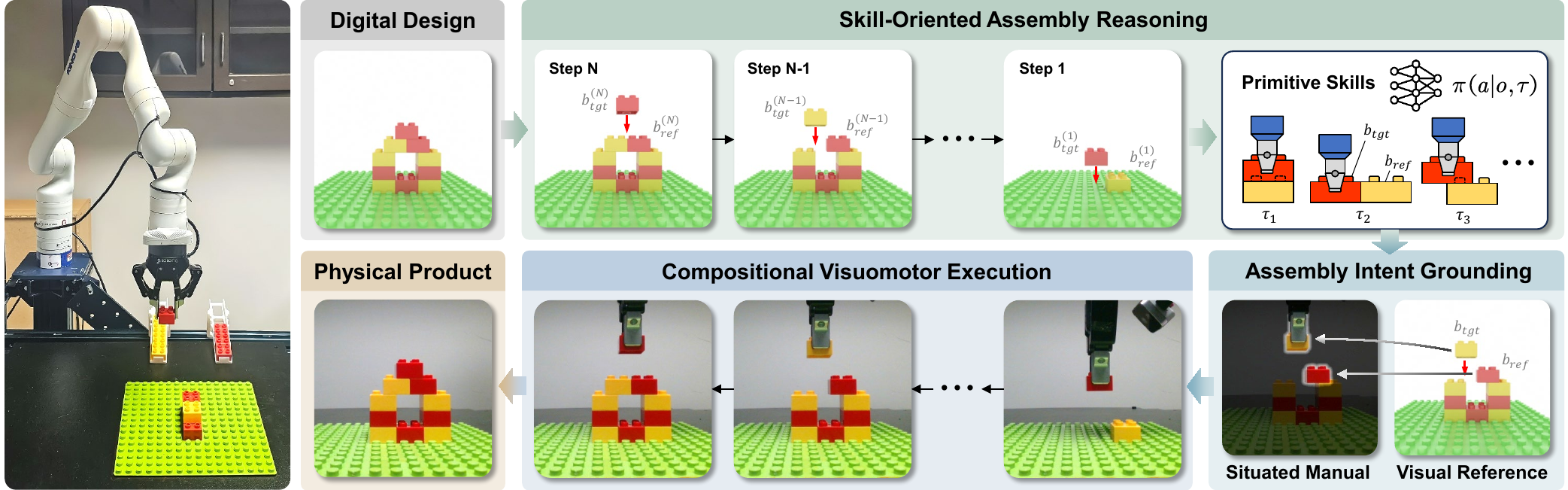}
	\captionof{figure}{\footnotesize \textbf{Overview of \textsc{BrickCraft}.} \textsc{BrickCraft} transforms a digital design into a physical product through three phases: (i) \textbf{Skill-Oriented Assembly Reasoning} decomposes the long-horizon task into steps anchored to reference bricks and maps them to reusable primitive skills. (ii) \textbf{Assembly Intent Grounding} generates situated manuals to provide spatial guidance; and (iii) \textbf{Compositional Visuomotor Execution} chains visuomotor skills to complete the assembly.}
    \label{figure1}
    \vspace{-5pt}
\end{strip}
\begin{abstract}
Autonomous robotic assembly of interlocking bricks demands seamless integration of long-horizon task reasoning, spatial grounding, and fine-grained manipulation. This paper presents \textsc{BrickCraft}, a compositional framework designed for long-horizon and generalizable interlocking brick assembly. \textsc{BrickCraft} models the assembly process using a relative formulation, where each step is anchored to a reference brick within the partial structure, thereby decomposing complex tasks into a finite set of reusable primitive skills. \textsc{BrickCraft} bridges the gap between high-level assembly plans and physical execution through situated manuals, which provide explicit spatial guidance for learned visuomotor skills by projecting the assembly intent onto real-time robot observations. Finally, \textsc{BrickCraft} employs a modular execution pipeline that chains these spatially grounded skills to accomplish long-horizon assembly tasks. Extensive experimental validations demonstrate that \textsc{BrickCraft} acquires proficient assembly skills from a limited set of demonstrations and effectively composes them for diverse long-horizon assembly tasks. The project website is available at \url{https://intelligent-control-lab.github.io/BrickCraft}. 
\end{abstract}

\section{INTRODUCTION}

Autonomous long-horizon assembly remains a formidable open problem in robotics, as it necessitates the seamless integration of physically grounded task reasoning and dexterous, contact-rich manipulation \cite{liu2025prompt,tian2025fabrica}. Taking interlocking brick assembly as an example, a robot must reason about structural dependencies to generate a feasible assembly sequence while executing fine-grained actions under tight geometric constraints. Mastering such capabilities represents a fundamental step toward intelligent embodied systems in complex real-world settings. While recent advances in learning-based approaches have demonstrated remarkable proficiency in dexterous manipulation and impressive generalization \cite{zitkovich2023rt, black2025pi05}, monolithic end-to-end control paradigms remain challenging for long-horizon assembly due to the inherent difficulty of implicitly modeling long-term dependencies and extensive data requirements. 

Hierarchical frameworks offer a viable alternative by decoupling task reasoning from execution \cite{tie2025manual2skill, xu2025query}. Analogous to generating and following an assembly manual, these approaches first employ a task planner to decompose the task into a sequence of actionable subgoals, and then invoke downstream learned skills for execution. However, a critical bottleneck lies in bridging high-level semantic reasoning and low-level visuomotor execution. Existing solutions typically address this gap using either explicit geometric states such as object poses \cite{sun2024arch}, or implicit representations such as goal images or multimodal embeddings \cite{gu2025manualvla}. The former relies on accurate perception and geometric estimation, making it sensitive to errors in tight-clearance assembly, while the latter provides only indirect spatial guidance and may be difficult to ground precisely in the physical scene. This challenge is particularly pronounced for interlocking bricks due to repetitive visual features and various possible interlocking configurations. We therefore seek a situated plan-to-execution interface to facilitate spatial grounding, alongside a modular formulation that maps diverse assembly configurations to reusable primitives.

To this end, we present \textsc{BrickCraft}, a framework that tackles long-horizon interlocking brick assembly by composing visuomotor skills with situated manual guidance. As illustrated in Fig. \ref{figure1}, given a 3D digital design, \textsc{BrickCraft} first plans a feasible assembly sequence via an assembly-by-disassembly strategy \cite{tian2022assemble}. Each assembly step is anchored to a specific reference brick, forming a relative representation that abstracts open-ended assembly tasks into a finite set of reusable primitive skills. To translate the symbolic plan into spatially grounded execution, we introduce situated manuals, a strategy that segments and highlights task-relevant entities directly within real-time robot observations to provide explicit visual guidance for learned visuomotor skills. In this way, \textsc{BrickCraft} integrates the complementary strengths of symbolic planning and learned visuomotor control, preserving task-level interpretability and compositionality while enabling robust closed-loop execution without requiring precise calibration or localization. Empirical evaluations show that situated manual guidance effectively resolves visual ambiguities and facilitates generalization to unseen brick structures, enabling the visuomotor skills to achieve an 86.25\% success rate across 240 assembly trials on 48 distinct structures. Furthermore, these skills can be reliably composed to accomplish various long-horizon assembly tasks.

Our main contributions are summarized as follows:
\begin{itemize}
    \item We present \textsc{BrickCraft}, a compositional framework for long-horizon interlocking brick assembly.
    \item We propose a novel task formulation based on relative interlocking relationships that maps diverse assembly tasks to a finite set of primitive skills.
    \item We propose situated manuals, a strategy that grounds symbolic intents into real-time robot observations to provide explicit spatial guidance for visuomotor execution.
    \item We validate \textsc{BrickCraft} through comprehensive real-world experiments across diverse single-step and long-horizon assembly tasks.
\end{itemize}

\section{RELATED WORK}
\textbf{Robotic Assembly of Interlocking Bricks}:
Interlocking brick assembly serves as an ideal long-horizon manipulation benchmark due to its universally accessible, standardized bricks and inherent combinatorial reconfigurability \cite{wen2026bricksim}, where a limited set of elementary components can be assembled into diverse structures. Prior studies have approached robotic brick assembly through task and motion planning \cite{huang2025apex}, relying on carefully engineered setups to achieve the high-precision brick pose estimation and execution necessary for tight-clearance insertion \cite{barghi2024bricks, liu2024lightweight, tang2025eye}. Alternatively, reinforcement learning \cite{fan2019learning} has been explored to alleviate the reliance on precise spatial alignment, which remains restricted to isolated, single-step insertions. ManualVLA \cite{gu2025manualvla} employs Chain-of-Thought reasoning to perform intermediate-goal prediction and assembly execution in an end-to-end manner, while the demonstrated assembly horizon and task diversity remain relatively limited. In contrast, BrickCraft decouples task planning from execution and composes reusable learned skills for long-horizon assembly, improving data efficiency and adaptation to novel tasks.

\textbf{Visuomotor Policies and Visual Guidance}: Generative visuomotor policies such as ACT \cite{zhao2023ACT} and diffusion policy \cite{chi2025diffusion, reuss2023goal} model conditional action distributions over multimodal observations and have demonstrated remarkable efficacy in task-specific, fine-grained manipulation. Vision-Language-Action (VLA) models \cite{black2025pi05, zitkovich2023rt} further leverage large-scale multimodal pre-training and language conditioning to achieve broader semantic understanding and cross-task generalization. However, end-to-end policies must implicitly infer spatial relationships from raw visual observations, which can hinder robust spatial generalization under limited demonstrations. Recent works therefore introduce explicit visual guidance to enhance spatial awareness. AimBot \cite{dai2025aimbot} overlays auxiliary visual cues onto RGB observations to explicitly encode end-effector states and spatial relationships with scene objects, while Gaze2Act \cite{zuo2026gaze2act} grounds human gaze into object masks and gaze points for target specification and enhanced visual attention. In BrickCraft, explicit visual guidance is instead derived from symbolic assembly plans to ground step-specific assembly intent and support skill generalization across different structures.

\textbf{Skill Composition for Long-Horizon Manipulation}:
Skill-centric frameworks achieve generalizable long-horizon manipulation by decomposing complex tasks into reusable primitive skills \cite{ sun2024arch, tie2025manual2skill, liu2025prompt}, where each skill can either be a learned policy or a motion planning primitive. MOSAIC \cite{mishani2025mosaic} composes heterogeneous skills through Generator and Connector modules to form executable skill sequences. Skill Graph \cite{yu2026autonomous} organizes semantically defined skills and feasible transitions into a graph for structured composition and execution. BrickCraft builds upon the skill-centric paradigm, while continuously aligning physical execution with the symbolic assembly plan through situated-manual grounding and step-wise state verification, thereby facilitating robust long-horizon assembly.

\section{METHOD}

\subsection{Problem Formulation and Framework Overview}
Given a target brick structure, \textsc{BrickCraft} completes long-horizon assembly in the physical world through three phases: (i) \textbf{Skill-Oriented Assembly Reasoning}, which infers a physically feasible assembly plan and maps each step to a specific primitive skill; (ii) \textbf{Assembly Intent Grounding}, which projects abstract assembly intents onto robot observations, forming situated manuals to establish spatial guidance; and (iii) \textbf{Compositional Visuomotor Execution}, which performs the assembly plan by composing reusable visuomotor skills under the guidance of situated manuals. The specific notations and formulations are detailed below.

\textbf{Assembly Task}: We consider the task of assembling a given multi-brick structure, denoted as $\mathcal{B} = \{b_1,b_2,\ldots,b_N\}$, on a planar baseplate. We restrict our scope to rectangular bricks with standardized stud spacing. Consequently, each brick $b_i$ is parameterized as $b_i = (c_i, p_i, \theta_i)$, with $c_i$ denoting the brick type, $p_i \in \mathbb{Z}^3$ denoting the brick position within a stud-based discrete grid system, and $\theta_i\in\{0,1,2,3\}$ denoting the discrete planar orientation index, which corresponds to the rotation angles of $0, \frac{\pi}{2}, \pi,$ and $\frac{3\pi}{2}$, respectively.

\begin{figure} 
	\centering
	\setlength{\fboxrule}{1pt}
	\setlength{\fboxsep}{2pt}
	\includegraphics[width=\dimexpr\columnwidth-2\fboxrule-2\fboxsep\relax]{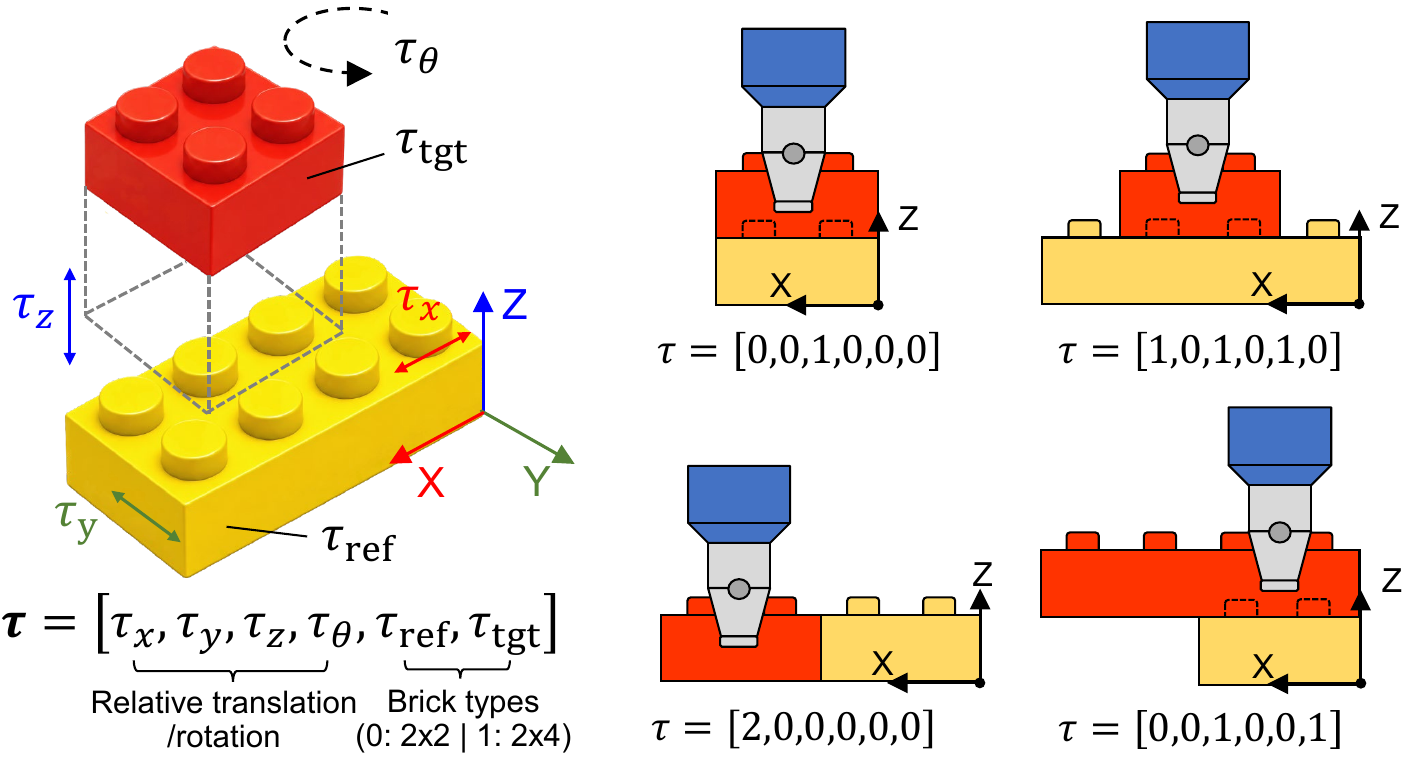}
	\caption{\footnotesize \textbf{Task encoding.} The 6D integer vector $\boldsymbol{\tau}$ encodes the relative translation and rotation between the target brick $b_{\mathrm{tgt}}$ and the reference brick $b_{\mathrm{ref}}$, as well as their respective brick types.}
    \label{figure2}
    \vspace{-10pt}
\end{figure}

\textbf{Assembly Plan}: The assembly plan is formulated as an ordered sequence $\mathcal{P} = \{\eta^{(1)}, \eta^{(2)}, \ldots, \eta^{(N)}\}$, where each assembly step $\eta = (b_{ref}, b_{tgt}, \boldsymbol{\tau})$ describes the action of attaching a target brick $b_{tgt}$ onto a reference brick $b_{ref}$ within the existing partial assembly. This action is parameterized by a geometric task encoding $\boldsymbol{\tau} = [\tau_x, \tau_y, \tau_z, \tau_\theta, \tau_{ref}, \tau_{tgt}] \in \mathcal{T}$, which encodes both the relative attachment pose and the brick types of $b_{ref}$ and $b_{tgt}$. As illustrated in Fig. \ref{figure2}, the translation components $[\tau_x, \tau_y, \tau_z]$ denote the discrete spatial offsets measured in stud units, while the rotation component $\tau_\theta \in\{0,1,2,3\}$ specifies the relative planar orientation. $\tau_{ref}$ and $\tau_{tgt}$ indicate the types of the reference and target bricks, respectively. This representation explicitly encodes the spatial multi-modality of brick connections, enabling the abstraction of diverse brick assembly tasks into a finite set of reusable primitive skills that can be effectively learned and generalized across novel structures.

\textbf{Visuomotor Skill}: A visuomotor skill $S: \mathcal{O} \to \mathcal{A}$ is defined as a parameterized primitive that maps an observation $o \in \mathcal{O}$ to an action $a \in \mathcal{A}$. In this work, the observation $o = (s, I_{ws}, I_{wrist})$ comprises the robot proprioceptive state $s$ and visual inputs $I_{ws}, I_{wrist} \in \mathbb{R}^{H \times W \times 3}$ from a workspace camera and a wrist camera, respectively.
For the brick assembly task, we consider two types of skills: pick skills $S_{pick}$ and assembly skills $S_{assembly}$. The pick skill is a modular grasping primitive for acquiring the target brick prior to assembly. We primarily focus on the assembly skill, which is formulated as a learned visuomotor policy $\pi(a \mid o, \boldsymbol{\tau})$. The task encoding $\boldsymbol{\tau} \in \mathcal{T}$ serves as a unique skill descriptor, where each $\boldsymbol{\tau}$ corresponds to a distinct primitive assembly skill. This formulation unifies diverse assembly behaviors within a single policy, facilitating shared representations across skills.

\textbf{Situated Manual}: To bridge the abstract assembly plan with the physical environment, we first render the expected state of each assembly step $\eta$ in the Blender simulation engine \cite{blender} to generate a visual reference $\mathcal{I}_r$ that incorporates necessary spatial and semantic cues. Given the current workspace observation $I_{ws}$, a spatial mapping $\mathcal{G}: (\mathcal{I}_r, I_{ws}) \rightarrow \tilde{I}_{ws}$ is formulated to project the assembly intent onto the robot observation. The resulting intent-augmented observation $\tilde{I}_{ws}$ is defined as the \textit{situated manual}, which serves as the input to the assembly skill to provide step-wise spatial guidance.

\begin{figure*}[!t]
	\centering
    \begingroup
    \setlength{\fboxrule}{1pt}
    \setlength{\fboxsep}{2pt}
    \includegraphics[width=\dimexpr\linewidth-2\fboxrule-2\fboxsep\relax]{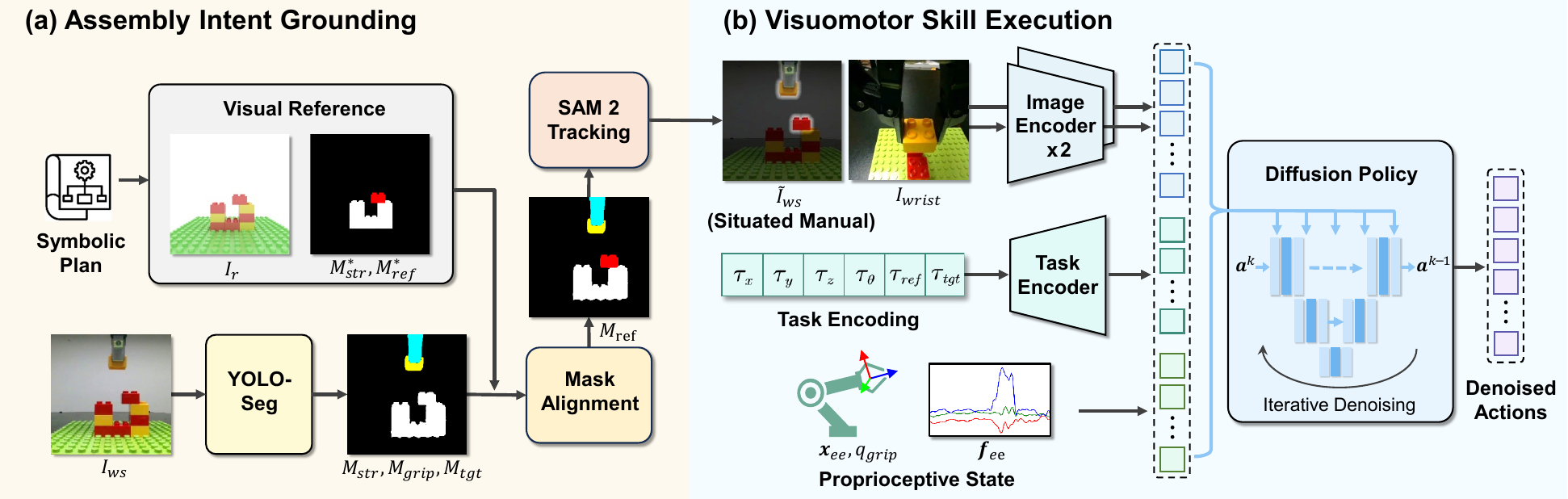}
    \endgroup
	\caption{\footnotesize \textbf{Situated manual-guided visuomotor assembly.} (a) \textbf{Assembly Intent Grounding}: Symbolic assembly plans are rendered into visual references in simulation and aligned with real-world observations $I_{ws}$ to extract task-relevant entity masks. These masks are tracked via SAM 2 \cite{ravi2024sam2} and overlaid onto real-time observations to yield the situated manual. (b) \textbf{Visuomotor Skill Execution}: We formulate the assembly skill as a diffusion policy \cite{chi2025diffusion}. The policy takes the situated manual as observation input and is conditioned on the task encoding $\boldsymbol{\tau}$ to generate diverse assembly behaviors.}
    \label{figure3}
    \vspace{-10pt}
\end{figure*}
\subsection{Skill-Oriented Assembly Reasoning}
Given a target structure $\mathcal{B}$ and a predefined set of task encodings $\mathcal{T}$, we employ an assembly-by-disassembly strategy \cite{liu2025prompt} to decompose the long-horizon assembly task. Specifically, starting from the goal state, we utilize a Depth-First Search (DFS) algorithm to explore viable disassembly paths. At each search step, the algorithm recursively identifies valid $(b_{ref}, b_{tgt}, \boldsymbol{\tau})$ tuples to iteratively deconstruct the structure until no bricks remain. We then reverse the obtained disassembly sequence to derive the forward assembly plan $\mathcal{P}$.

During the search process, we evaluate whether a candidate brick $\hat{b}$ is removable from the current partial structure $\hat{\mathcal{B}}$ under the task encoding $\boldsymbol{\tau} \in \mathcal{T}$ according to the following criteria:

\textbf{1) Connectivity} ($\mathcal{C}_{con}$): There exists a valid reference brick $\hat{b}_{ref} \in \hat{\mathcal{B}} \setminus \{\hat{b}\}$ such that the relative spatial relationship between $\hat{b}$ and $\hat{b}_{ref}$ matches the task encoding $\boldsymbol{\tau}$.

\textbf{2) Observability} ($\mathcal{C}_{obs}$): The target brick $\hat{b}$ and its reference brick $\hat{b}_{ref}$ maintain unoccluded lines of sight to both the workspace and wrist cameras.

\textbf{3) Operability} ($\mathcal{C}_{op}$): The brick $\hat{b}$ can be removed without collisions. Specifically, assuming a vertical disassembly trajectory along the $z$-axis, the swept volume generated by both the brick and the robotic gripper must remain collision-free with respect to the remaining structure $\hat{\mathcal{B}} \setminus \{\hat{b}\}$.

\textbf{4) Stability} ($\mathcal{C}_{stab}$): The remaining structure is physically stable. We leverage StableLego \cite{liu2024stablelego} to evaluate the structural stability.

A removal action is considered valid only if it satisfies all four criteria, such that $\mathcal{C} = \mathcal{C}_{con} \land \mathcal{C}_{obs} \land \mathcal{C}_{op} \land \mathcal{C}_{stab} = 1$.

\subsection{Assembly Intent Grounding as Situated Manuals}\label{section:manual}
Given the symbolic assembly plan $\mathcal{P}$, the next step is to translate these task-level instructions into actionable guidance for the downstream visuomotor policy.  The visuomotor policy employs complementary dual-view observations: the workspace camera provides a side view of the assembled structure, whereas the wrist camera provides a close-range, downward-looking view centered on the manipulated brick. However, the periodic geometric units and highly homogeneous visual features of the assembled structure make it difficult for the visuomotor policy to distinguish the intended assembly location. To mitigate this issue, we propose grounding the assembly intent in the workspace observation as a situated manual to provide explicit spatial guidance. Essentially, this situated manual serves as an attentional augmentation of the raw camera observation, where a spotlight is applied to the reference brick to anchor the visual focus of the robot. We apply this augmentation only to the workspace camera, as the wrist view is inherently task-centered, with its narrow field of view naturally focusing on the local assembly interface. The overall pipeline for this grounding process is depicted in Fig. \ref{figure3}(a), which comprises four primary steps:

\textbf{1) Visual Reference Generation}: Assuming the CAD models of the bricks are known, we utilize the Blender simulation engine \cite{blender} to render the visual references. For each assembly step $\eta$, the simulation engine renders the current state of the assembly to virtually reconstruct the cumulative assembly process. The resulting visual reference $\mathcal{I}_{r}$ encapsulates a photorealistic RGB rendering $I_{r}$ and two complementary semantic masks: a global mask $M^*_{str}$ covering the entire existing assembly structure, and a local mask $M^*_{ref}$ isolating the specific reference brick $b_{ref}$. This representation provides the necessary spatial cues for the robot to align its physical observation with the prescribed virtual state.

\textbf{2) Physical Entity Segmentation}: Before the execution of each assembly skill, we employ a fine-tuned YOLOv8-seg model \cite{yolov8} to extract semantic entities from the raw workspace camera observation $I_{ws}$. Specifically, this segmentation module isolates three critical functional components within the physical scene: the current assembled structure on the baseplate, the robotic gripper, and the target brick currently held by the gripper. The model outputs a corresponding set of physical masks $\{M_{str}, M_{grip}, M_{tgt}\}$. This step isolates task-relevant components from the background, facilitating the subsequent alignment process.

\textbf{3) Mask Alignment}: To locate the reference brick in the physical workspace, we perform a geometric alignment between the camera observation and the visual reference. Under the assumption that the simulated camera is coarsely aligned with the physical viewpoint, we employ the enhanced correlation coefficient (ECC) algorithm \cite{evangelidis2008ECC} to estimate an optimal affine transformation matrix $H$ that maps the rendered structure mask $M^*_{str}$ to its observed counterpart $M_{str}$. By applying this transformation, the virtual reference mask $M_{ref}^*$ is projected into the physical observation to yield $M_{ref}$. 

\textbf{4) Mask Tracking and Overlay}: The task-relevant masks identify the reference and target bricks at the initialization of each assembly step. As the target brick moves during execution, we employ Segment Anything Model 2 (SAM 2) \cite{ravi2024sam2} to continuously track these masks, with its memory architecture maintaining temporal consistency under transient occlusions. In each control cycle, these masks are projected onto the raw observation $I_{ws}$ via a background dimming strategy, which creates a spotlight-like effect on the target entities. Specifically, the brightness of regions outside the active masks is attenuated to $25\%$ of their original intensity, with a $10$-pixel gradient transition applied at the mask boundaries to avoid sharp artificial edges. This visual augmentation, yielding a high-contrast situated manual, highlights task-relevant entities and suppresses distractor features within the observation, effectively preventing visual ambiguity in complex structural configurations and providing precise assembly guidance for the robot.

\begin{figure*}[!t]
	\centering
    \includegraphics[width=0.95\linewidth]{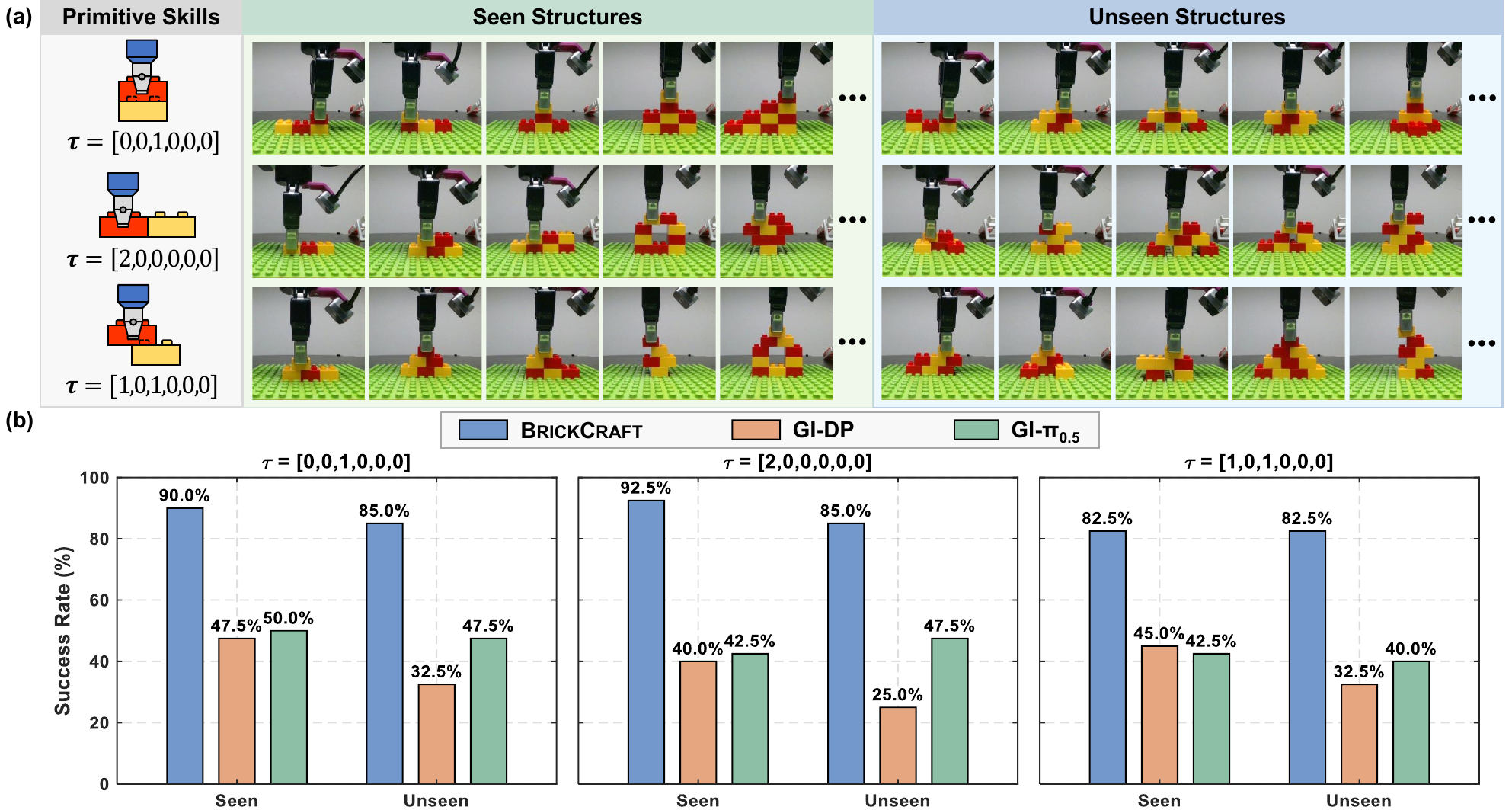}
	\caption{\footnotesize \textbf{Evaluation of visuomotor assembly skills.} We evaluate three distinct primitive skills, testing each skill across 8 seen and 8 unseen structural configurations, with 5 independent trials per structure. (a) Demonstrations of visuomotor assembly on diverse structures; (b) Success rate comparison (40 trials per bar). \textsc{BrickCraft} consistently outperforms the baselines, demonstrating robust generalization to unseen structures.}
    \label{exp_skill}
    \vspace{-10pt}
\end{figure*}
\subsection{Visuomotor Skill Learning}

We parameterize the visuomotor assembly skill using a Diffusion Policy \cite{chi2025diffusion}, motivated by its strong capability in capturing multi-modal action distributions. As illustrated in Fig. \ref{figure3}(b), the conditional input comprises dual-view visual observations, proprioceptive state and task encoding, where the workspace camera view is augmented to serve as the situated manual. The visual observations are encoded via separate ResNet-18 backbones, and the resulting visual features are concatenated. The task encoding is mapped into a 32-dimensional embedding space via a task encoder consisting of an embedding layer followed by an MLP. The proprioceptive state includes the end-effector pose $\bm{x}_{ee}$, gripper state $q_{grip}$, and 3D end-effector force $\bm{f}_{ee}$ estimated from joint torques. We employ a CNN-based U-Net as the noise prediction head, which is conditioned on the concatenated features to generate the action chunks.

To train the visuomotor skills, we collect expert demonstrations across structurally diverse brick assembly tasks. These demonstrations are subsequently categorized into individual skill-based datasets. The visual observations are further processed using the assembly intent grounding pipeline introduced in Sec. \ref{section:manual} to incorporate the requisite task-relevant spatial cues for effective skill learning.

\subsection{Skill Composition for Long-Horizon Assembly}\label{section:skill_composition}

Long-horizon assembly is executed by composing visuomotor skills according to the assembly plan $\mathcal{P}$. Specifically, each assembly step $\eta^{(i)} = (b_{ref}^{(i)}, b_{tgt}^{(i)}, \boldsymbol{\tau}^{(i)}) \in \mathcal{P}$ is realized through a skill pair $\{S_{pick}^{(i)}, S_{assembly}^{(i)}\}$. The pick skill $S_{pick}^{(i)}$ is conditioned on the target brick type $c_{tgt}^{(i)}$ to guide visual localization and grasping, while the assembly skill $S_{assembly}^{(i)}$ is conditioned on the task encoding $\boldsymbol{\tau}^{(i)}$ to execute the designated assembly configuration. The long-horizon task is then executed as a skill chain: $\{S^{(1)}_{pick} \to S^{(1)}_{assembly} \to S^{(2)}_{pick} \to \cdots \to S^{(N)}_{assembly} \}$. At each step, \(S_{\mathrm{assembly}}^{(i)}\) is activated from a coarse pre-assembly pose within the training distribution and performs closed-loop visuomotor alignment and insertion without relying on a precise target pose.

To regulate transitions along this skill chain, each skill is associated with a pre-condition \(\mathcal{C}_{pre}(o)\) for skill initiation and a post-condition \(\mathcal{C}_{post}(o)\) for outcome verification. For the pick skill, $\mathcal{C}_{pre}$ verifies target visibility and kinematic feasibility, while $\mathcal{C}_{post}$ confirms a stable grasp. For the assembly skill, $\mathcal{C}_{pre}$ verifies the consistency between the observed structure mask $M_{str}^{(i)}$ and the rendered current-state mask $M_{str}^{*,(i)}$ after alignment. $\mathcal{C}_{post}$ evaluates the assembly outcome by comparing the expected post-assembly mask $\hat{M}_{str}^{(i+1)}$, obtained from the rendered visual reference of the next assembly state, with the newly observed structure mask $M_{str}^{(i+1)}$ based on their Intersection over Union (IoU). The assembly is considered successful when the IoU exceeds a predefined threshold. By aligning real-world observations with the symbolic assembly plan at each step, the system can identify execution errors and mitigate compounding drift. The state verification further enables the system to autonomously identify the current assembly state and resume execution from the corresponding step following an interruption.


\section{EXPERIMENT}
In this section, we conduct comprehensive experiments to address the following core research questions:

(1) Can the proposed situated manual effectively guide the visuomotor skills to execute high-precision assemblies and generalize to unseen brick structures? 

(2) Can BrickCraft effectively compose visuomotor skills to accomplish long-horizon assembly?

\subsection{Experimental Setup and Implementation Details}

\textbf{System Setup}: 
The hardware platform employs a Kinova Gen3 robotic arm equipped with a Robotiq 2F-85 gripper. Visual observations are captured by two Intel RealSense cameras: one mounted on the robot wrist and another mounted on the table side. The assembly tasks are performed using LEGO Duplo bricks of two shapes (2x2 and 2x4) and three colors. The bricks are sorted by color and shape and placed on supply rails for picking at predefined locations, allowing the evaluation to focus on the assembly stage.

\textbf{Data Collection, Training and Inference Details}:
We collected 865 demonstration trajectories for 7 assembly skills, each corresponding to a distinct task encoding, across more than 80 assembly configurations. The observations and paired actions were sampled at 10 Hz. Visual inputs were resized to 256x256 and randomly cropped to 224x224 during training. We first manually annotated approximately 50 representative images to fine-tune the YOLOv8-seg and SAM 2 models. The full demonstration dataset was then augmented using the assembly intent grounding pipeline and used to train the diffusion-based assembly skills. For the diffusion process, we employed the DDIM scheduler with 100 denoising iterations during training and 10 iterations during inference. During assembly execution, action commands are published at 10 Hz. YOLOv8-seg is invoked only once at the beginning of each assembly skill to initialize SAM 2 tracking. On a workstation equipped with an NVIDIA RTX 4090 GPU, the average inference latencies of SAM 2 and the diffusion policy are 15.6 ms and 49.2 ms, respectively.

\textbf{Baselines}:  To evaluate how situated manual guidance benefits visuomotor skill learning and generalization, we compare BrickCraft against the following baselines: (1) \textbf{Goal Image-Conditioned Diffusion Policy (GI-DP)} \cite{reuss2023goal}, which takes the goal image of the current assembly step as an additional conditioning input, without situated manual guidance or task encoding. For each demonstration, we generate a rendered goal image matching the workspace-camera viewpoint. The model is trained on a mixed dataset of rendered and real goal images, which we found helps the image encoder learn more robust visual representations. During evaluation, the baseline is conditioned solely on rendered goal images for consistency with BrickCraft. (2) \textbf{Goal Image-Conditioned $\pi_{0.5}$ (GI-$\pi_{0.5}$)} \cite{black2025pi05}, a state-of-the-art VLA baseline fine-tuned from the official pre-trained weights using the same goal image-conditioned dataset.

\textbf{Evaluation Metrics}: For single-step assembly, the \textbf{success rate} evaluates the performance of visuomotor assembly skills. An assembly step is considered successful if the target brick is correctly installed without deforming or detaching previously assembled bricks. For long-horizon evaluation, the \textbf{cumulative success rate} at step $k$ is the proportion of trials that complete the first $k$ steps without failure, characterizing execution reliability as the task horizon increases. Additionally, we report the \textbf{completion rate}, defined as the ratio of successfully completed assembly steps to the total number of steps in a task, to quantify the overall progress of execution.

\subsection{Results and Discussion}
\begin{figure}[!t]
	\centering
    \includegraphics[width=\columnwidth]{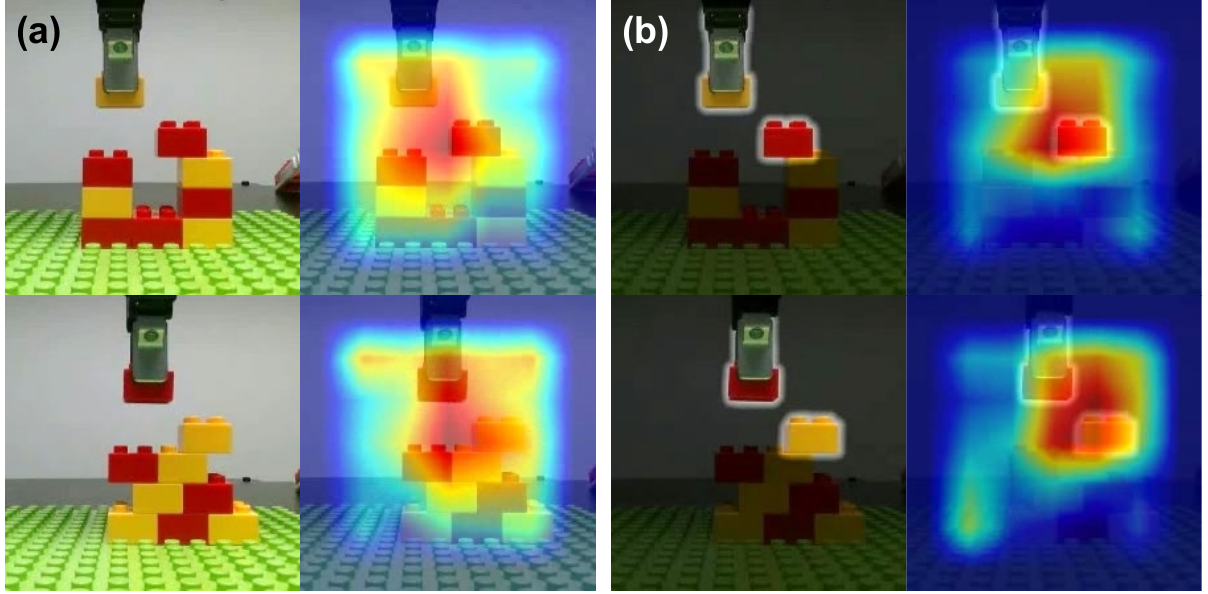}
	\caption{\footnotesize EigenCAM \cite{muhammad2020eigen} heatmaps for (a) the GI-DP baseline and (b) \textsc{BrickCraft} with situated manual guidance.}
    \label{CAM_vis}
    \vspace{-10pt}
\end{figure}
\begin{figure}[!t]
	\centering
    \begingroup
    \setlength{\fboxrule}{1pt}
    \setlength{\fboxsep}{2pt}
    \includegraphics[width=\dimexpr1.0\columnwidth-2\fboxrule-2\fboxsep\relax]{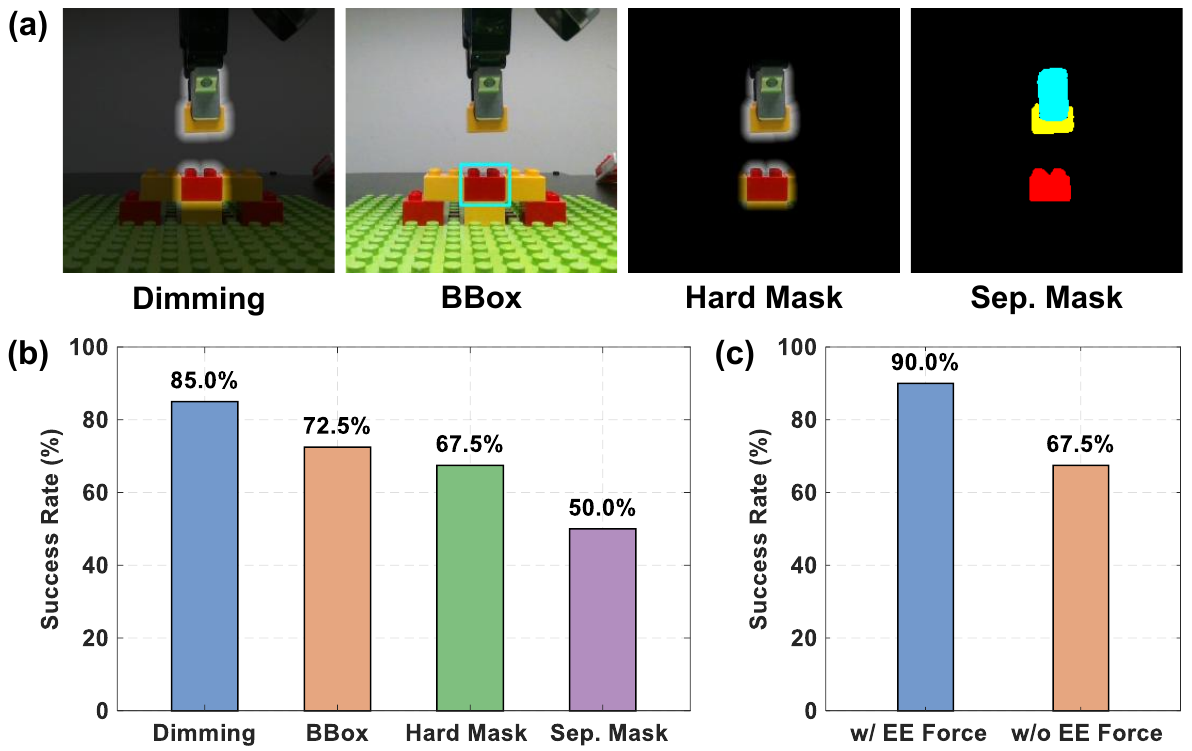}
    \endgroup
	\caption{\footnotesize \textbf{Ablation studies.} (a) Examples of different situated manual representations: background dimming (Dimming), bounding-box (BBox), hard masking (Hard Mask), and separate mask input (Sep. Mask). (b) Success rate comparison of different situated manual representations. (c) Success rate comparison of proprioceptive states with and without end-effector force observation. }
    \label{exp_ablation}
    \vspace{-10pt}
\end{figure}
\textbf{Skill Performance}: 
We evaluate the single-step assembly performance across three primitive skills, as illustrated in Fig. \ref{exp_skill}(a). Each skill and the corresponding baselines are trained on skill-specific sub-datasets (approximately 200 demonstrations per skill) and evaluated on 8 seen and 8 unseen structures, with 5 randomized trials per structure. Results are summarized in Fig.~\ref{exp_skill}(b). \textsc{BrickCraft} achieves an overall success rate of 86.25\% across 240 trials. Comparable performance is observed on seen and unseen structures, indicating robust structural generalization. Both GI-DP and GI-$\pi_{0.5}$ achieve success rates below 50\%, reflecting the difficulty of learning fine-grained geometric representations from raw visual inputs with limited demonstrations, as required for precise assembly. GI-DP further exhibits a 12.5\% to 15.0\% drop on unseen structures, whereas GI-$\pi_{0.5}$ maintains relatively stable performance across seen and unseen settings, likely due to the stronger semantic representations acquired through extensive multimodal pre-training.

To gain deeper insights into how situated manual guidance informs the visuomotor skill, we employ EigenCAM \cite{muhammad2020eigen} to visualize the principal components of the visual feature map for GI-DP and \textsc{BrickCraft}. As illustrated in Fig. \ref{CAM_vis}, the feature activations of GI-DP are notably dispersed, rendering the policy vulnerable to task-irrelevant distractors in the environment. Conversely, the situated manual isolates the geometric invariants across diverse assembly tasks, directing the visual attention of the policy toward the relative geometric relationship between the target and reference bricks. This concentrated spatial grounding minimizes visual ambiguity, thereby facilitating efficient skill learning and robust generalization to novel structures.

\begin{figure*}[!t]
    \centering
    \begingroup
    \setlength{\fboxrule}{1pt}
    \setlength{\fboxsep}{2pt}
    \includegraphics[width=\dimexpr\linewidth-2\fboxrule-2\fboxsep\relax]{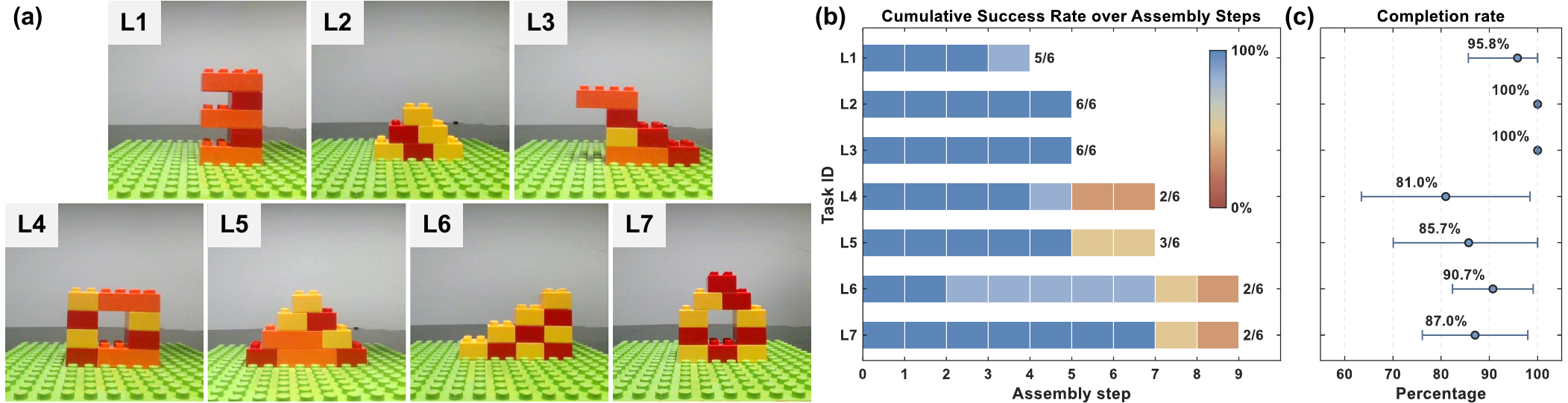}
    \endgroup
    \caption{\footnotesize \textbf{Evaluation of long-horizon assembly.} (a) Seven assembly tasks (L1–L7) with diverse shapes and brick counts. (b) Cumulative success rate over assembly steps. Bar length denotes the task horizon; annotations indicate the number of full-task successes. (c) Average completion rate; error bars indicate the standard deviation. Each task is evaluated over six trials.}
    \label{exp_long_horizon}
    \vspace{-10pt}
\end{figure*}
\textbf{Ablation on Situated Manual Representation}: To assess our representational choice for the situated manual, we conduct an ablation study comparing four ways of incorporating task-relevant masks into the visual observation, as shown in Fig.~\ref{exp_ablation}(a). Specifically, Dimming attenuates the non-masked image regions to 25\% intensity; BBox highlights the reference brick with a bounding box; Hard Mask removes the background and retains only the masked RGB regions; and Sep. Mask provides the raw RGB observation together with a separate color-coded mask. We evaluate the four representations on 8 unseen tasks, with 40 trials in total. As reported in Fig.~\ref{exp_ablation}(b), BBox and Hard Mask underperform Dimming by 12.5 and 17.5 percentage points, respectively. We speculate that BBox may be susceptible to background distractions, whereas Hard Mask may be more sensitive to slight mask inaccuracies. Sep. Mask achieves the lowest success rate, possibly because the policy struggles to associate the separate mask with the corresponding regions in the RGB observation. Beyond the specific choice of prompting representation, the results suggest that injecting semantic intent into the visual observations provides an effective and practical interface for spatial grounding and conditioning visuomotor skills.

\textbf{Ablation on End-Effector Force Observation}: To assess the contribution of end-effector force observation, we trained two policies with and without the 3D end-effector force input under identical training settings and evaluated them on 8 seen assembly tasks with 40 trials in total. Removing the force input reduces the success rate from 90.0\% to 67.5\%, as shown in Fig.~\ref{exp_ablation}(c).  We observed that the force-free policy occasionally released the gripper before sufficient contact force was established, resulting in incomplete interlocking. This suggests that end-effector force provides complementary contact-state cues that facilitate action prediction during contact-rich assembly.

\begin{figure}[!t]
	\centering
    \begingroup
    \setlength{\fboxrule}{1pt}
    \setlength{\fboxsep}{2pt}
    \includegraphics[width=\dimexpr\columnwidth-2\fboxrule-2\fboxsep\relax]{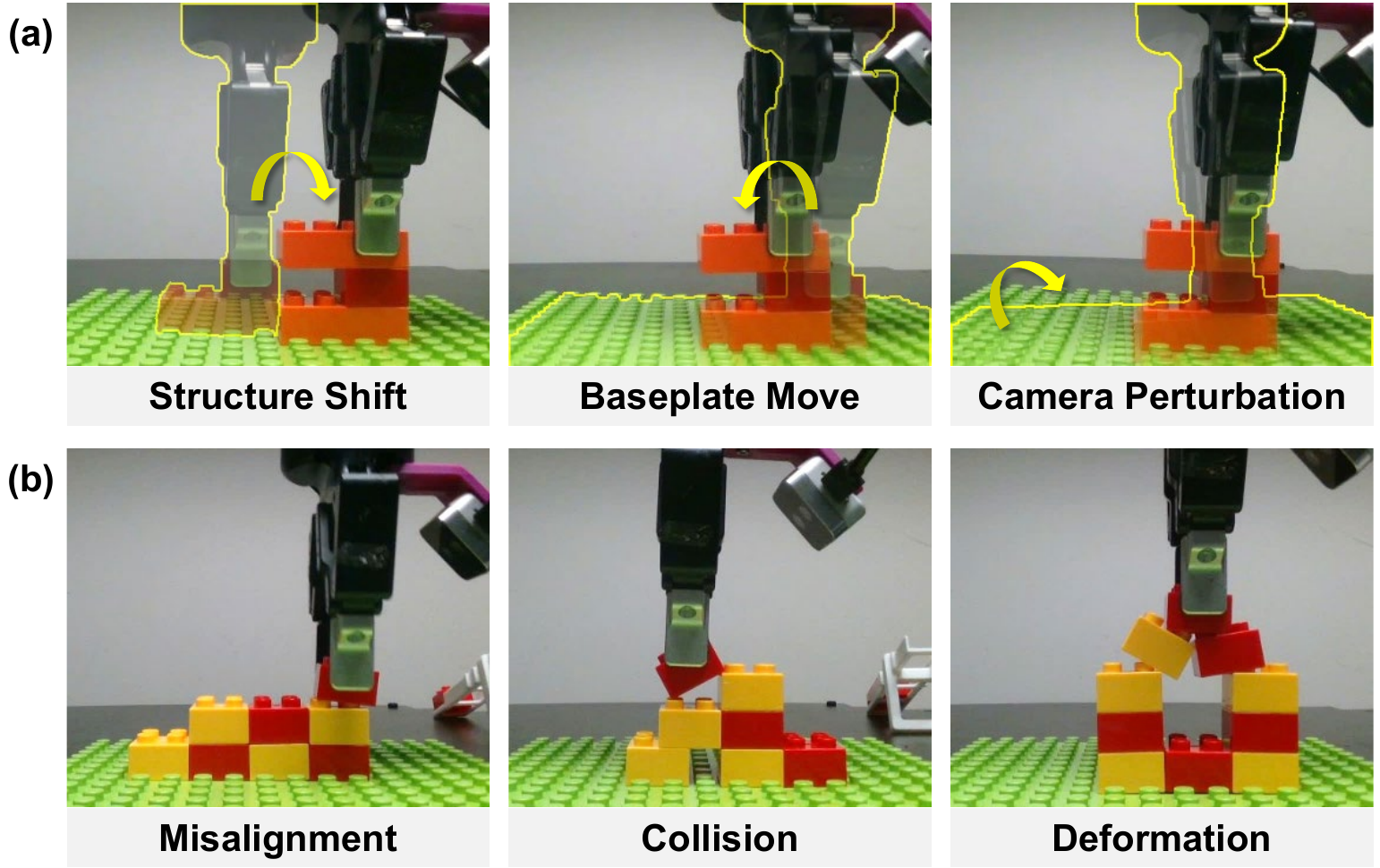}
    \endgroup
	\caption{\footnotesize (a) Robustness under spatial perturbations. The semi-transparent overlay depicts the previous-step assembly pose. (b) Typical failure modes.}
    \label{exp_perturb}
    \vspace{-10pt}
\end{figure}
\textbf{Long-Horizon Assembly Performance}: We conduct long-horizon assembly evaluations across seven tasks (L1--L7) with diverse structure shapes and task horizons, as depicted in Fig. \ref{exp_long_horizon}(a). For each task, \textsc{BrickCraft} autonomously generates an assembly plan from the target structure. The initial brick is manually placed to serve as a spatial anchor, after which the remaining assembly is executed autonomously. We perform 6 independent trials per task. Successfully completed assembly steps took 15.5 s on average, including both the pick and assembly phases. As shown in Fig.~\ref{exp_long_horizon}(c), the average completion rates remain above 80\% for all seven tasks, demonstrating that \textsc{BrickCraft} can effectively compose reusable skills to accomplish diverse multi-step assembly tasks. Meanwhile, Fig.~\ref{exp_long_horizon}(b) shows the step-wise evolution of cumulative success rates for each task. Shorter tasks generally achieve higher full-task success, with performance declining as task horizons become longer. Failures tend to occur at later stages, as imperfect interlocking may accumulate across steps, making subsequent assembly progressively more challenging. L6 exhibits a relatively low cumulative success profile but a high completion rate, since cumulative success is determined by the first failure, whereas completion rate still counts successful subsequent steps when execution remains feasible.

We further qualitatively evaluate system robustness to spatial perturbations during long-horizon execution. As shown in Fig. \ref{exp_perturb}(a), we introduce perturbations by shifting the assembled structure, moving the baseplate, and changing the camera viewpoint, while the robot can still complete the subsequent assembly steps. Among all successful steps, the maximum translational offset from assembly-skill activation to final assembly was 222.3 mm, reflecting the spatial variation handled during execution. These results demonstrate that BrickCraft can adapt to environmental changes without relying on a calibrated or fixed workspace setup.

\textbf{Failure Cases Analysis}: Typical failure modes during brick assembly are shown in Fig.~\ref{exp_perturb}(b). \textit{Misalignment} occurs when downward pressure is applied before sufficiently alignment, causing contact interference that prevents successful interlocking. \textit{Collision} may occur when the learned policy generates suboptimal motions that interfere with adjacent bricks. Incorporating human-in-the-loop correction to provide targeted corrective demonstrations for these failure states may further improve skill reliability. \textit{Structural deformation} is mainly observed in partially supported configurations, where insertion forces exceed the mechanical tolerance of the structure. Although end-effector force observations provide useful contact-state cues, simple concatenation with high-dimensional visual features may introduce modality imbalance, making it difficult for the policy to capture fine-grained contact dynamics. Our future work will investigate more structured force--vision fusion mechanisms \cite{lee2025manipforce, kang2025robotic} to facilitate more precise force-aware adaptation.

\section{CONCLUSION}
We present \textsc{BrickCraft}, a compositional framework for long-horizon interlocking brick assembly. By modeling the assembly process through a relative formulation anchored to reference bricks, we abstract open-ended assembly space into a finite set of reusable primitive skills. We then introduce situated manuals to ground symbolic assembly intents into real-time visual observations to provide explicit spatial guidance for visuomotor execution. Extensive real-world experiments demonstrate that situated manual guidance effectively suppresses task-irrelevant visual distractors, facilitating efficient skill learning and robust generalization to unseen structures. Furthermore, by chaining these spatially grounded skills into a composable execution pipeline, \textsc{BrickCraft} autonomously accomplishes diverse long-horizon assembly tasks.

To further elevate the autonomy of \textsc{BrickCraft}, future work will investigate autonomous failure recovery across diverse failure modes, including adaptive strategies such as re-grasping and deformation correction. In particular, large multimodal foundation models could provide a promising basis for failure reasoning and recovery planning. Beyond interlocking brick assembly, we believe that the central idea underlying this work, namely bridging symbolic reasoning with spatially grounded atomic skills, offers a promising pathway toward deploying robust embodied agents in broader real-world manipulation scenarios.




\section*{APPENDIX}

\setcounter{figure}{0}
\renewcommand{\thefigure}{A\arabic{figure}}
\renewcommand{\theHfigure}{appendix.A.\arabic{figure}}
\setcounter{table}{0}
\renewcommand{\thetable}{A\arabic{table}}
\renewcommand{\theHtable}{appendix.A.\arabic{table}}

\subsection{Additional Implementation Details}

We implement the visuomotor assembly skills using a U-Net-based diffusion policy developed based on the LeRobot~\cite{cadenelerobot} framework. The corresponding hyperparameters are summarized in Table~\ref{diffusion_hyperparameters}. For SAM 2-based mask tracking, we use the \texttt{sam2\_hiera\_tiny} model with approximately 39M parameters and an input image size of $256\times256$.

\begin{figure}[!t]
	\centering
    \begingroup
    \setlength{\fboxrule}{1pt}
    \setlength{\fboxsep}{2pt}
    \includegraphics[width=\dimexpr\columnwidth-2\fboxrule-2\fboxsep\relax]{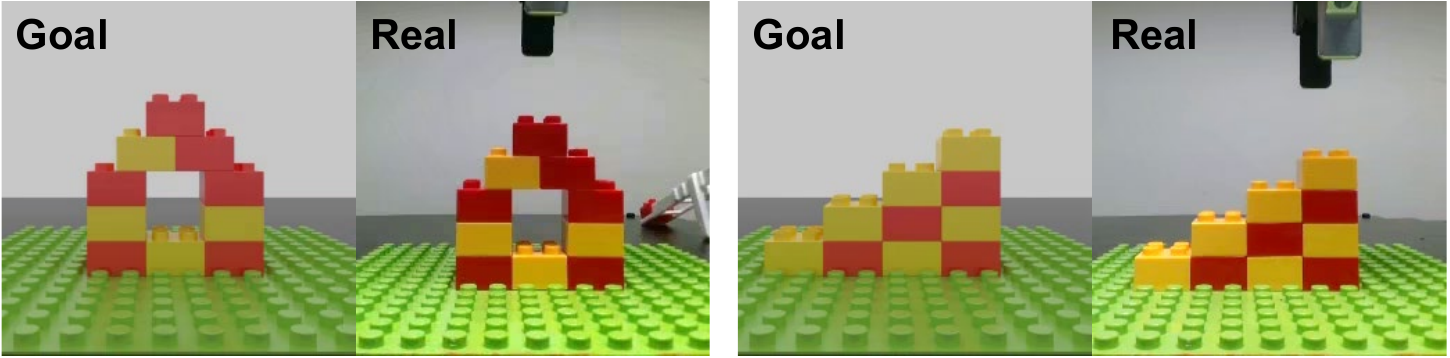}
    \endgroup
	\caption{\footnotesize Representative rendered goal images and corresponding real observations for the goal-image-conditioned baselines.}
    \label{goal_image}
    \vspace{-10pt}
\end{figure}

For the GI-DP baseline, we extend the same U-Net-based diffusion policy by introducing the goal image as an additional visual conditioning input, encoded by a separate ResNet-18 encoder. We use Blender \cite{blender} to render goal images that approximately match the workspace-camera viewpoint. Representative rendered goal images and their corresponding real observations are shown in Fig.~\ref{goal_image}. The goal image remains fixed throughout each episode. For GI-$\pi_{0.5}$, the image resolution, state and action dimensions, action prediction and execution horizons follow the settings in Table~\ref{diffusion_hyperparameters}.

\begin{center}
    \centering
    \captionof{table}{\footnotesize Diffusion policy hyperparameters for visuomotor assembly skills.}
    \label{diffusion_hyperparameters}
    \footnotesize
    \setlength{\tabcolsep}{3pt}
    \begin{tabular}{@{}ll@{}}
        \toprule
        \textbf{Hyperparameter} & \textbf{Value} \\
        \midrule
        Image resolution & $224 \times 224$ \\
        Proprioceptive state dimension & 13 ($\mathbf{x}_{ee}$: $9^{*}$ + $q_{\mathrm{grip}}$: 1 + $\mathbf{f}_{ee}$: 3) \\
        Action dimension & 10 ($\mathbf{x}_{ee}$: $9^{*}$ + $q_{\mathrm{grip}}$: 1) \\
        Action prediction horizon & 16 \\
        Action execution horizon & 6 \\
        Image encoder & ResNet-18 \\
        Image feature dimension & 64 per camera\\
        Task encoding dimension & 6 \\
        Task encoder & $6 \times 8$D embeddings + 2-layer MLP \\
        Task feature dimension & 32 \\
        Total parameters & 280M \\
        \bottomrule
        \multicolumn{2}{@{}l@{}}{\scriptsize \textsuperscript{*}We use the 6D rotation representation introduced in \cite{zhou2019continuity}.} \\
    \end{tabular}
\end{center}

\subsection{Post-Condition Checking for Assembly Verification}
\begin{figure}[H]
	\centering
    \begingroup
    \setlength{\fboxrule}{1pt}
    \setlength{\fboxsep}{2pt}
    \includegraphics[width=\dimexpr\columnwidth-2\fboxrule-2\fboxsep\relax]{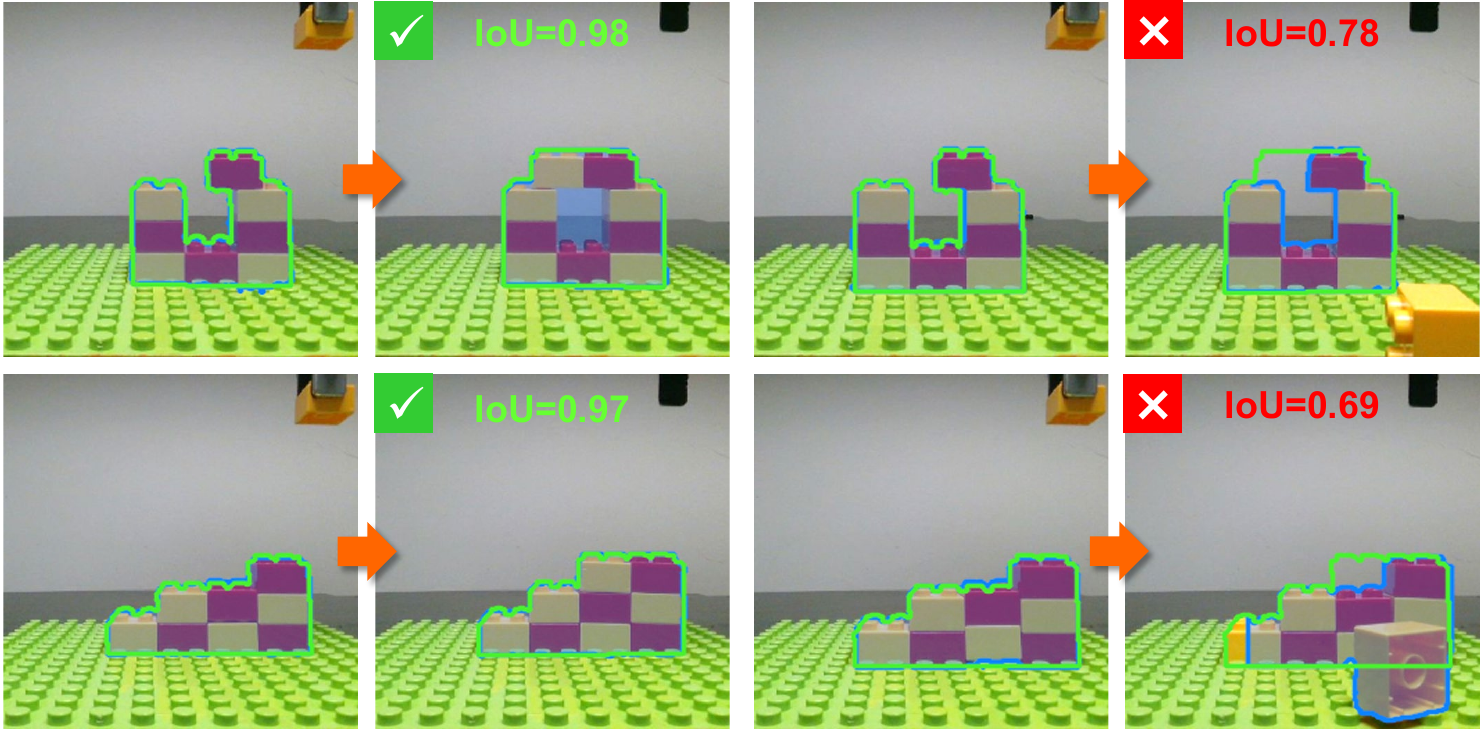}
    \endgroup
	\caption{\footnotesize Examples of assembly post-condition checking. Green and blue contours denote the expected and observed structure masks, respectively. A high IoU between the predicted and observed structure masks indicates successful assembly, whereas a low IoU indicates failure.}
    \label{fig:post_condition_examples}
    \vspace{-10pt}
\end{figure}

Fig.~\ref{fig:post_condition_examples} illustrates the assembly post-condition checking procedure described in Sec.~\ref{section:skill_composition}. After each assembly execution, the expected post-assembly mask from the rendered reference is compared with the observed structure mask based on their IoU, where the observed mask is obtained from the updated workspace-camera image via YOLOv8-seg. An assembly step is considered successfully completed when the IoU exceeds a predefined threshold.

\bibliographystyle{IEEEtran}
\bibliography{IEEEabrv, References}
\end{document}